\documentclass[10pt]{article}
\usepackage{styles/acl2015}
\usepackage{amscd,amsmath}
\usepackage{amsfonts}
\usepackage[round, numbers, authoryear]{natbib}
\usepackage{amssymb}
\usepackage{enumitem}
\usepackage{graphicx}
\usepackage{multirow}
\usepackage{longtable}
\usepackage{adjustbox}
\usepackage{float}
\usepackage{times}
\usepackage{url}
\usepackage{latexsym}
\usepackage{subfig}
\usepackage{array}
\usepackage{xcolor}
\usepackage[ruled,linesnumbered,vlined,algosection]{algorithm2e}
\title{An Evolutionary Algorithm for the Vehicle Routing Problem with Drones with Interceptions}

\author{Carlos Pambo\\
  Department of Industrial Engineering \\
  Faculty of Engineering \\
  Stellenbosch University \\
  {\tt carlos.pambo@hotmail.com} \\\And
  Jacomine Grobler \\
  Department of Industrial Engineering \\
  Faculty of Engineering \\
  Stellenbosch University \\
  {\tt jacomine.grobler@sun.ac.za} \\}

\date{}

\begin{document}
\maketitle
\begin{abstract}
 The use of trucks and drones as a solution to address \textit{last-mile} delivery challenges is a new and promising research direction explored in this paper. The variation of the problem where the drone can intercept the truck while in movement or at the customer location is part of an optimisation problem called the vehicle routing problem with drones with interception (VRPDi).
This paper proposes an evolutionary algorithm (EA) to solve the VRPDi. The paper demonstrates a metaheuristic strategy by applying an evolution-based algorithm to solve the VRPDi. In this variation of the VRPDi, multiple pairs of trucks and drones need to be scheduled. The pairs leave and return to a depot location together or separately to make deliveries to customer nodes. The drone can intercept the truck after the delivery or meet up with the truck at the following customer location.
The algorithm was executed on the travelling salesman problem with drones (TSPD) datasets by \cite{bouman2015datasets}, and the performance of the algorithm was compared by benchmarking the results of the VRPDi against the results of the VRP of the same dataset. This comparison showed improvements in total delivery time between $39\%$ and $60\%$. 
Further detailed analysis of the algorithm results examined the total delivery time, distance, node delivery scheduling and the degree of diversity during the algorithm execution. This analysis also considered how the algorithm handled the VRPDi constraints. The results of the algorithm were then benchmarked against algorithms in \cite{dillon2023investigating} and \cite{ernst2024phdthesis}. The latter solved the problem with a maximum drone distance constraint added to the VRPDi.
The analysis and benchmarking of the algorithm results showed that the algorithm satisfactorily solved $50$ and $100$-nodes problems in a reasonable amount of time, and the solutions found were better than those found by the algorithms in \cite{dillon2023investigating} and \cite{ernst2024phdthesis} for the same problems. However, the algorithm performance deteriorated considerably as the number of nodes in the problems increased. This deterioration was in terms of the quality of the solution and the computation time required to solve the problem.
\end{abstract}

\section{Introduction}

Over the past few years, there has been a massive adoption of online shopping. Several factors contributed to that reality, such as increased access to internet services, the rise in smartphone users, and companies increasingly looking to decrease the time between consumers ordering products and these orders being delivered. These factors have all contributed to the increase in the adoption of online shopping (e-commerce) as a standard method of shopping. This adoption has led to delivery companies expanding their delivery networks and looking for ways to optimise delivering goods from warehouses to customers. This last leg of the delivery network, known as last-mile delivery operations/scheduling, can be characterised as a vehicle routing problem (VRP) optimisation problem.

Adding drones to the VRP creates new flexibility and efficiency in the transportation system. Drones can deliver goods directly to customers, bypassing road congestion and thus reducing delivery times. They can also access remote or difficult-to-reach locations, such as islands or mountainous areas, where the accessibility for ground vehicles may be limited. Also, drones are considered a green form of transportation.
Solving the VRPDi represents a significant challenge due to the complexity of integrating drones into the existing VRP frameworks. It is a very active research area in engineering, transportation, operations research, and computer science. Researchers have explored various approaches, including heuristic algorithms \citep{daknama2017vehicle,schermer2018algorithms,di2021trucks,rios2021recent}; metaheuristics algorithms \citep{braekers2016vehicle,schermer2019matheuristic, pugliese2020using,rios2021recent,ernst2021algos}; and exact methods \citep{braekers2016vehicle, di2021trucks} to find efficient solutions to the VRPD. Developing efficient and effective VRPD algorithms can significantly improve the performance and cost-effectiveness of \textit{last-mile} delivery operations.

\textit{Problem.} The VRPDi is a variant of the well-known VRP that includes using drones with interception capabilities, that is, return to the vehicle from where they are launched after the delivery. In contrast, the vehicle is stationary or in transit to the next delivery customer to support transporting goods. The VRP can be categorised as a combinatorial optimisation problem in operations research, where the aim is to minimise an objective function. This objective function could be the total distance travelled, the total delivery time required, and the total cost of the scheduling operation. In the instance of VRPDi, the distance travelled by the drones from the vehicle where it is launched to the customer location and its return to the launching vehicle also counts towards the problem's total cost, subject to various constraints such as time windows and the capacity of the vehicles.

\textit{Approach.} This paper develops an evolutionary algorithm (EA) to address the vehicle routing problem with drones with interception (VRPDi), where the use of drones is included to support the transportation of goods, where a combination of vehicles and drones can deliver goods, where the drone can intercept the vehicle mid-way through its next delivery node. VRPDi introduces new complexities to the VRP, requiring coordination and synchronisation between vehicles to optimise the routing and scheduling of both vehicles and drones, as well as the time and energy needed for intercepting and transferring packages from the vehicle to the drone.

\textit{Contribution and structure.} Our contribution is the first use of a genetic algorithm (GA) to solve the VRPDi.The approach includes defining the number of trucks and drone combinations and using a metameric representation of the algorithm solution. A review of the current VRPDi literature is presented in Section \ref{sectionLiteratureReview}, including the mathematical formulation of the problem in this paper and related work. Section \ref{sectionApproach} overviews our approach, including a detailed description of the algorithm developed. Section \ref{sectionResults} discusses the results of the algorithm using the different benchmark problems. Section \ref{sectionDiscussion} presents a discussion of the results obtained by the algorithm, while Section \ref{sectionConclusions} concludes the paper.

\section{Background} \label{sectionLiteratureReview}

\begin{figure}[H] 
    \centering
    \includegraphics[width=.5\textwidth]{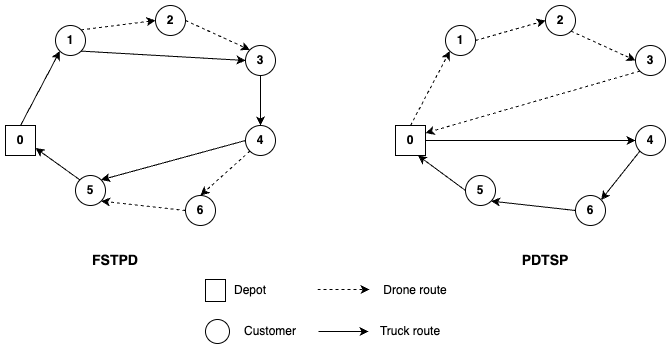}
    \caption[FSTSP and PDSTP \cite{murray2015flying}]{FSTSP and PDSTP \cite{murray2015flying}.} 
    \label{figFSTP}
\end{figure}

The introduction of drones in collaboration with trucks to perform delivery tasks resulted in the creation of the truck drone routing problem (TDRP), which is a variation of the classical VRP,  which was first introduced in \cite{dantzig1959truck} and has been an intensively researched over the decades since. Contrasting both VRP and TDRP is the introduction of drones in the latter and the inherited need for the solution to simultaneously cater to the collaboration between drones and trucks. Early research on the TDRP aimed to address solutions for the travelling salesman problem with drones (TSPD), a simplified version of the TDRP consisting of only one truck.
In \cite{murray2015flying}, two classes of the TSP were studied: the flying sidekick travelling salesman problem (FSTSP) and the parallel drone scheduling travelling salesman problem (PDSTSP). Both problems assume that a truck or a drone serves a customer. In the FSTSP, a drone collaborates with a truck to serve different customers, is dispatched from the truck to make deliveries and is collected by the truck to restore resources such as batteries and cargo. In the PDSTSP, the truck and drone work independently to serve the customers. Figure \ref{figFSTP} illustrates both problems.
A variation of TDRP, which has multiple trucks, referred to as a vehicle routing problem with drones (VRPD), was first introduced by \cite{poikonen2017vehicle}. VRPD consists of a fleet of trucks carrying a certain number of drones to deliver packages to customer nodes.
\cite{wang2017vehicle} researched using multiple trucks and drones to deliver packages to customers in a VRPD scenario. Each customer could only request one package at a time and could only be visited once. The trucks and drones began and ended their journeys at the depot, and the drone could be launched either from the depot or from the truck. The truck and drone could only meet at the truck's next customer's location, and the truck that launched the drone had to retrieve it. Whichever vehicle arrived at the customer's location first waited for the other.
The focus of this paper is on the VRPD, studied in \cite{wang2017vehicle, poikonen2017vehicle}. In this variation of the problem, a drone is launched from the truck at a customer node and re-joins the same truck it was launched from at the next customer node to be visited by the truck. The VRPD study by \cite{moremi2022ant} introduces the ability of the truck to intercept the drone at an intermediate point after the drone has completed its delivery.
Different variants of the VRPD exist in the literature. These variants consider different combinations of settings for the truck-drone pair and objectives for the system. Objectives include minimising energy costs, tour time, delivery cost, number of drones in the system, and total travel costs. Different solution methods have addressed these objectives \citep{wang2017vehicle, wang2019vehicle, schermer2019hybrid, sacramento2019adaptive}.

The VRPDi in this study is defined in a graph $\textbf{G}= (\textbf{N}, \textbf{A})$. Where $\textbf{N}$ represents the node-set, containing the depot node and a set of customer nodes $\textbf{l} = \{ {l}_{1}, {l}_{2}, \ldots, {l}_{n-1}, {l}_{n} \}$, and $\textbf{A} = \{( i, j ) | i, j \in N, i \neq j\}$ represents the arc set. For convenience in notation, ${o}^{s}$ and ${o}^{t}$, represent the origin and termination depot, respectively. The problem contains a fleet of $m$ homogeneous trucks, each carrying one drone. The trucks travel at a constant speed of ${v}_{t}$ and take ${\omega}_{i}$ time to perform the delivery. The drones travel at a constant speed of ${v}_{d}$ and take ${\sigma}_{i}$ time to perform the delivery. The truck and the drone are initially located at the origin depot, ready to serve the customers. For simplicity, this paper does not adopt a docking or transfer hub to land drones, which is preferred in the drone delivery industry \citep{wang2019vehicle}. The station primarily intends to store and maintain backup and support drones for landing purposes. During pilot experiments, drones can take off from any location, but they can only land at designated docking hubs or depots, not at the customer's location. This is because drone landings require specific conditions, such as ample space and a particular docking device that can communicate with the drone to ensure a precise and secure landing. Due to the above conditions for drone landing not being easily reproducible at customer locations, in addition to safety and privacy concerns, this paper opts for drone deliveries to be performed via airdrop instead of landing at the customer node. A truck can be loaded up to its maximum capacity ($C$) units of customer parcels. Each truck in the system can carry a maximum of one drone. Each drone in the system can fly from the truck carrying it located at any of the locations in $\textbf{N}$, with a customer parcel.

The system does not impose the maximum flying duration of a drone as ${T}^{D}$, meaning that there is no restriction in terms of from which node a drone can fly to perform a delivery as long that specific node has not been visited by a truck or a drone previously. Every time the drone returns from delivery, it is fitted with new batteries, and the next parcel is loaded. These operations are very fast and, therefore, have no impact on the total cost.

\begin{figure}[H] 
    \centering
    \includegraphics[width=0.5\textwidth, height=6cm]{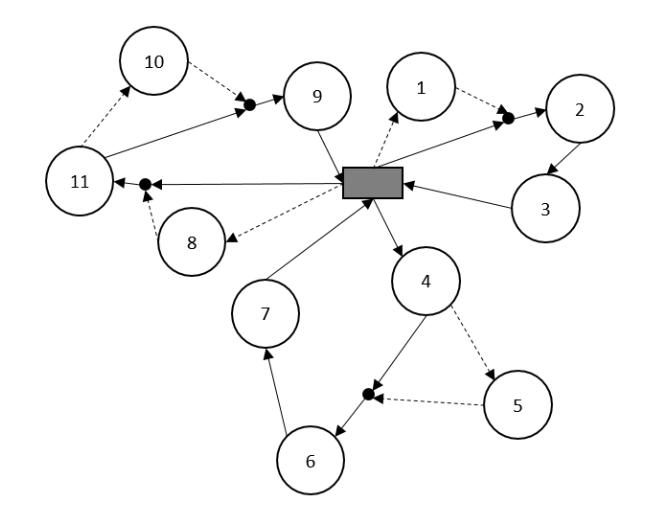}
    \caption[Vehicle routing problem with drones with interceptions]{Vehicle routing problem with drones with interceptions} 
    \label{figVRPDi}
\end{figure}

It's worth noting that in case a truck finishes its current delivery, and has to drive to the next customer location it will visit, while its drone is also performing a delivery at another node, the truck has two options, taking its speed and drone speed, its current location and the drone location into account. The first is to calculate the time it will require to drive to the position of the possible intermediate interception location, between its current location and its next customer node location, plus the waiting time for the drone to arrive. The second option is to calculate the time it will require to drive to its next customer location, perform the delivery and wait for the drone to arrive from its delivery at that customer location. Whichever of the options requires less time becomes the next action the truck will perform. The mathematical formulation of the above is explained in detail in the next section.
The objective of this VRPDi is to minimise the total delivery time. Total delivery time is represented by the maximum amount of time taken by the slowest of the truck-drone pair to finalise all of its customer deliveries. This delivery time of the truck-drone pair is represented by the sum of the time required by the truck and the drone to perform their deliveries.
Figure \ref{figVRPDi} depicts the VRPDi. The example problem consists of three truck-drone pairs, one depot, and eleven nodes to be serviced. The first truck visits nodes 2 and 3, the second one visits nodes 4, 6, and 7, and the third one visits nodes 11 and 9. The first drone visits node 1, the second drone visits node 5, and the third drone visits node 10. Interception points are black circles, solid lines represent truck routes, and dotted lines represent drone routes. VRPDi has forbidden moves such as launching a drone at an intermediate location or only allowing one drone delivery between truck deliveries. The drone can only fly back to the truck from which it was launched.

\subsection{Mathematical Formulation} \label{sectionVRPDiMathematicalFormulation}

\begin{figure}[H] 
    \centering
    \includegraphics[width=.5\textwidth]{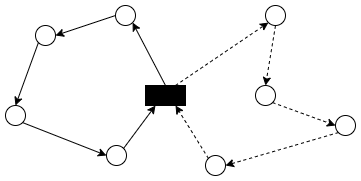}
    \caption[Example of a VRP routing solution]{Example of a VRP routing solution}
    \label{figVRPRoutingSolution}
\end{figure}

The vehicle routing problem (VRP) seeks to find optimal delivery or collection routes from a depot to customers while considering constraints. The problem was first introduced as the truck dispatching problem (TDP) by \cite{dantzig1959truck}. While exact methods are used to solve VRP optimally, heuristic methods like the Clark and Wright savings algorithm are used to solve large VRPs \citep{clarke1964scheduling}. Over time, VRPs became more complex, resulting in different VRP variants. Examples of these different variations include the vehicle routing problem with time windows (VRPTW) \citep{desrochers1992new}, the vehicle routing problem with pick-ups and deliveries (VRPPD) \citep{min1989multiple}, and the multi-depot vehicle routing problem \citep{wren1972computer}.
The capacitated vehicle routing problem (CVRP), or traditional VRP, involves vehicles with limited capacities $(C)$ servicing customers scattered across a geographic area from a central depot. Figure \ref{figVRPRoutingSolution} shows a CVRP example. The black square represents the depot, while the circle symbols are customer nodes spread across an $x-y$ Cartesian plane. The dashed and solid lines denote the routes of the two vehicles. VRPs aim to determine the minimum distance or time to service all customer nodes while ensuring that each delivery route does not exceed the vehicle's capacity. When the vehicles' capacities differ, the problem becomes a varying fleet VRP.
The VRP is defined on a graph ($\textbf{G}$), where $\textbf{G} = (\textbf{V}, \textbf{A})$, and $\textbf{V} = \{0,...,n\}$. Node 0 is the depot, and each customer location $j$ has a known demand ${d}_{j}$.

Equation \ref{eqVRPPRoblemDemand} represents a set of edges. Each edge connecting nodes $i$ and $j$ has a non-negative cost ${c}_{ij}$ associated with it, which is the cost of travelling from node $i$ to node $j$. The asymmetric VRP has directed arcs ($\textbf{A}$), while the symmetric VRP has undirected arcs ($\textbf{E}$). Each customer in a set $\textbf{S}$ has a demand $d(\textbf{S})$ that equals the sum of all customers' demands in $\textbf{S}$, denoted by ${d}_{j}$. $K$ vehicles have a capacity of $C$. The variable ${x}_{ij}$ is a binary variable equal to $1$ if edge $(i, j)$ is serviced by a vehicle in the solution. Otherwise, ${x}_{ij}$ is $0$.
\begin{equation}
    \label{eqVRPPRoblemDemand}
    \begin{split}
        \textbf{A} = \{(i, j) : i, j \in \textbf{V}\}, \:\:\:\:\:\:\:\:\:\:\:\: i \neq j
    \end{split}
\end{equation}
For illustration, Equations \ref{eqVRPProblemEquation} to \ref{eqVRPProblemConstraint6} describe a two-index vehicle flow formulation of an asymmetrical VRP with specified assumptions \citep{laporte1992vehicle}:
\begin{itemize}
    \item Every vehicle's route begins and concludes at the depot.
    \item Every customer node, denoted as $j$ in the set $\textbf{V}$ except for node $0$, is visited only once.
    \item The total demand of customers visited by each vehicle should not exceed its capacity, denoted by $C$.
\end{itemize}
\begin{equation} 
    \label{eqVRPProblemEquation}
    \begin{split}
       \text{Minimise Z} = \sum^{n}_{i < j} {c}_{ij} {x}_{ij}
    \end{split}
\end{equation}
Subject to:
\begin{equation} 
    \label{eqVRPProblemConstraint1}
    \begin{split}
        \sum_{i \in \textbf{V} \setminus \{j\}} {x}_{ij} = 1  \:\:\:\:\:\: {\forall} {j} \in \textbf{V}\setminus\{0\}
     \end{split}
\end{equation}
\begin{equation} 
    \label{eqVRPProblemConstraint2}
    \begin{split}
        \sum_{j \in \textbf{V} \setminus \{i\}} {x}_{ij} = 1  \:\:\:\:\:\: {\forall} {i} \in \textbf{V} \setminus \{0\}
     \end{split}
\end{equation}
\begin{equation} 
    \label{eqVRPProblemConstraint3}
    \begin{split}
        \sum_{i \in \textbf{V} \setminus \{i\}} {x}_{i0} = K
     \end{split}
\end{equation}
\begin{equation} 
    \label{eqVRPProblemConstraint4}
    \begin{split}
        \sum_{j \in \textbf{V} \setminus \{j\}} {x}_{0j} = K
     \end{split}
\end{equation}
\begin{equation} 
    \label{eqVRPProblemConstraint5}
    \begin{split}
        \sum_{i \notin \textbf{S}} \sum_{j \in \textbf{S}} {x}_{ij} \geq r(\textbf{S}) \:\:\:\:\:\: {\forall} {\textbf{S}} \subseteq \textbf{V} \setminus \{0\}, \textbf{ S} \neq \phi 
     \end{split}
\end{equation}
\begin{equation} 
    \label{eqVRPProblemConstraint6}
    \begin{split}
        {x}_{ij} \in {0, \text{ } 1} \:\:\:\:\:\: {\forall} {i, j} \in \textbf{V}
     \end{split}
\end{equation}
\begin{itemize}
    \item The goal of the function labelled as \ref{eqVRPProblemEquation} is to minimise the distance or travel time for each vehicle.
    \item Constraints \ref{eqVRPProblemConstraint1} and \ref{eqVRPProblemConstraint2} are constraints that ensure each vertex has one arc entering and leaving it.
    \item The requirements for the depot vertex's degree are outlined in constraints \ref{eqVRPProblemConstraint3} and \ref{eqVRPProblemConstraint4}.
    \item Constraints \ref{eqVRPProblemConstraint5} are constraints that eliminate subtours. $r(\textbf{S})$ is the lower bound on the number of vehicles needed to visit all vertices in $\textbf{S}$.
\end{itemize}
In \cite{moremi2022ant}, the vehicle routing problem with drone with interception (VRPDi), was introduced with the following characteristics:

To determine the time and position of the drone and the truck, Equation \ref{eqInterceptionPosition} was utilised. The equation calculates the interception time. This equation calculates the time taken by the drone to move from node $j$ to the truck's location $\textbf{{pI}}^{j}$. The vector $\textbf{{c}}^{j}$ is a 2-dimensional representation of the $x$ and $y$ coordinates of node $j$, whereas ${d}(\textbf{{c}}^{i}, \textbf{{c}}^{j})$ is the distance between node $j$ and the position of a truck at the beginning of a drone launch at node $i$.

To calculate the time from node $j$ to the interception point $\textbf{{pI}}^{j}$, $\textbf{{TI}}^{j}$ is used, and it is calculated in Equation \ref{eqIntTime}, where $\textbf{{c}}^{i} - \textbf{{c}}^{j}$ represents the difference between the $x$ and $y$ coordinates of nodes $j$ and $i$ as a 2-dimensional vector.

\begin{equation} 
    \label{eqIntTime}
    \resizebox{.9\hsize}{!}{$
        \textbf{{TI}}^{j} = \frac{-2(\textbf{{c}}^{i} - \textbf{{c}}^{j}).({v}^{ik}_{t}) \pm \sqrt{(2((\textbf{{c}}^{i} - \textbf{{c}}^{j}).(\textbf{{v}}^{ik}_{t}))^2-(4({v}^{2}_{d} - {v}^{2}_{t})(-d{(\textbf{{c}}^i, \textbf{{c}}^{j})}^2)))}}{2({v}^{2}_{d} - {v}^{2}_{t})}
    $}
\end{equation}

\begin{figure}[H] 
    \centering
    \includegraphics[width=.4\textwidth]{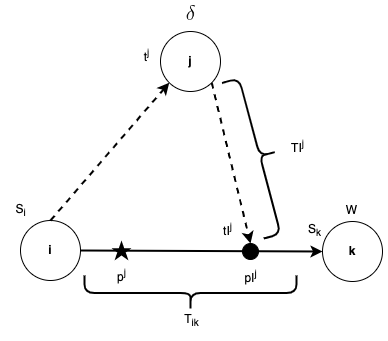}
    \caption[Truck-drone subtour \cite{moremi2022ant}]{Truck-drone subtour \cite{moremi2022ant}.} 
    \label{figTruckDroneSubtour}
\end{figure}
The interception position can then be calculated as follows:
\begin{equation} 
    \label{eqInterceptionPosition}
    \begin{split}
        \textbf{{pI}}^{j} = \textbf{{c}}^{i} + (\textbf{{v}}^{ik}_{t}).({TI}^{j} + \sigma + ({T}^{j} - {s}_{i}))
    \end{split}
\end{equation}

In this situation, the drone has two choices: intercept the truck before it gets to node $k$ or fly straight to node $k$. If the drone cannot intercept the truck before it reaches node $k$, it will fly directly to the node's location, and the truck will wait for the drone. The total time it takes for the truck to travel from node $i$ to $k$ is known as ${T}_{ik}$. The total time it takes for the truck to travel from node $i$ to the interception point $\textbf{{pI}}^{j}$ is equal to $({tI}^{j} - {s}_{i})$, where ${tI}^{j}$ is the time when the drone and the truck meet after the drone has delivered at node $j$ and ${s}_{i}$ is the time when the truck arrives at customer $i$ for delivery. This time is equivalent to the drone's travel time from $i$ to the interception point $\textbf{{pI}}^{j}$. If the truck arrives at node location $k$ without intercepting the drone, it must wait for its arrival. The waiting time, $W$, is calculated as ${T}_{ijk} - {T}_{ik}$, where ${T}_{ijk}$ is the total drone travel time from node $i$ to $k$ and is only relevant when ${t}_{ijk} > {T}_{ik}$. The total time to finish the deliveries of nodes $i$ and $j$ and travel to $k$ is ${T}_{ijk} = max({T}_{ik}, {t}_{ijk})$. The arrival of the last vehicle at node $k$ determines ${T}_{ijk}$.

\cite{moremi2022ant} proposed the following mathematical models with the assumption that a drone delivery must be scheduled between each truck delivery.The model has the following input parameters:
\begin{itemize}
    \item $n$ denotes the number of nodes to be serviced, including a depot at node 0
    \item ${T}_{ijk}$ denotes the time required to attend to nodes $i$ and $j$, and travel to $k$. The truck moves between nodes $i$ and $k$, while the drone caters to node $j$ in between.
\end{itemize}
In the above model, the decision variables are:
\begin{itemize}
    \item The variable ${u}_{i}$ is employed to remove sub-tours.
\end{itemize}
\begin{equation*} 
    {x}_{ijk} = \begin{cases}
                    1, \text{ If the truck services node } i \text{ before node } k, \\ \text{ and the drone services node } j \text{ in between} \: \\ 
                    0, \text{ Otherwise} 
                \end{cases}
\end{equation*}
\begin{equation*}
    {b}_{j} = \begin{cases}
                1, \text{ If the drone services node } j \\
                0, \text{ If the truck services node } j
            \end{cases}
\end{equation*}
\begin{equation*} 
    {z}_{ik} = \begin{cases}
        1, \text{ If the truck travels from node } i \text{ to node } k \\ 
        0, \text{ Otherwise}
    \end{cases}
\end{equation*}
\begin{equation} 
    \label{eqTSPDi1ObjectiveFunction}
    \begin{split}
        \text{\textit{min}}\sum_{i=1}^{n} \sum_{k=1}^{n} \sum_{j=1}^{n} {x}_{ijk} {T}_{ijk}
    \end{split}
\end{equation}
Subject to:
\begin{equation} 
    \label{eqTSPDi1Constrain1}
    \begin{split}
        {z}_{ii} = 0 \:\:\:\:\:\: {\forall} {i} \in \{1,...,n\}
    \end{split}
\end{equation}
\begin{equation} 
    \label{eqTSPDi1Constrain2}
    \begin{split}
        {x}_{iii} = 0 \:\:\:\:\:\: {\forall} {i} \in \{1,...,n\}
    \end{split}
\end{equation}
\begin{equation} 
    \label{eqTSPDi1Constrain3}
    \begin{split}
        {z}_{ik} = \sum_{j=1}^{n} {x}_{ijk} \:\:\:\:\:\: {\forall} {i}, {k} \in \{1,...,n\}
    \end{split}
\end{equation}
\begin{equation} 
    \label{eqTSPDi1Constrain4}
    \begin{split}
        {b}_{j} = \sum_{i=1}^{n}\sum_{k=1}^{n} {x}_{ijk} \:\:\:\:\:\: {\forall} {j} \in \{1,...,n\}
    \end{split}
\end{equation}
\begin{equation} 
    \label{eqTSPDi1Constrain5}
    \begin{split}
        \sum_{i=1}^{n} {z}_{iq} - \sum_{k=1}^{n} {z}_{qk} = 0 \:\:\:\:\:\: {\forall} {q} \in \{1,...,n\}
    \end{split}
\end{equation}
\begin{equation} 
    \label{eqTSPDi1Constrain6}
    \begin{split}
        \sum_{k=1}^{n} {z}_{ik} = 1 - {b}_{i} \:\:\:\:\:\: {\forall} {i} \in \{1,...,n\}
    \end{split}
\end{equation}
\begin{equation} 
    \label{eqTSPDi1Constrain7}
    \begin{split}
        \sum_{k=1}^{n} {z}_{ik} = 1 - {b}_{k} \:\:\:\:\:\: {\forall} {k} \in \{1,...,n\}
    \end{split}
\end{equation}
\begin{equation} 
    \label{eqTSPDi1Constrain8}
    \begin{split}
        \sum_{j=1}^{n} {x}_{ijk} \le 1 \:\:\:\:\:\: {\forall} {i},{k} \in \{1,...,n\}
    \end{split}
\end{equation}
\begin{equation} 
    \label{eqTSPDi1Constrain9}
    \begin{split}
       {u}_{i} + {z}_{ij} \le {u}_{j} + (n - 1)(1 - {z}_{ij}) \\ {\forall} {i},{j} \neq 1 \in \{1,...,n\}
    \end{split}
\end{equation}
\begin{equation} 
    \label{eqTSPDi1Constrain10}
    \begin{split}
        {u}_{1} = 0
    \end{split}
\end{equation}
\begin{equation} 
    \label{eqTSPDi1Constrain11}
    \begin{split}
       {x}_{ijk}, {b}_{i}, {z}_{ik} \in \{0, 1\} \:\:\:\:\:\: {x} {i}, {j}, {k}  \in \{1,...,n\}
    \end{split}
\end{equation}
\begin{itemize}
    \item The objective of Equation \ref{eqTSPDi1ObjectiveFunction} is to reduce the drone's and truck's overall travel time.
    \item Constraints \ref{eqTSPDi1Constrain1} and \ref{eqTSPDi1Constrain2} simplify the model by ensuring that the truck and drone cannot return to a previously visited node.
    \item Constraint \ref{eqTSPDi1Constrain3} links ${z}_{ik}$ and ${x}_{ijk}$.
    \item Constraint \ref{eqTSPDi1Constrain4} links ${b}_{j}$ and ${x}_{ijk}$.
    \item Constraint \ref{eqTSPDi1Constrain5} ensures that if a delivery is made at a node, the vehicle used for delivery must depart from that same node.
    \item Constraint \ref{eqTSPDi1Constrain8} ensures that only one drone delivery can occur between two truck deliveries.
    \item Constraints \ref{eqTSPDi1Constrain9} and \ref{eqTSPDi1Constrain10} eliminate any sub-tours.
    \item Constraint \ref{eqTSPDi1Constrain11} restricts the decision variables to binary values.
\end{itemize}

\section{Our approach} \label{sectionApproach}

Algorithm \ref{algoEAforVRPDi} provides a pseudo-code description of the algorithm used to solve the VRPDi. The algorithm requires the following input control parameters: the number of nodes ($n$) and the number of iterations the algorithm runs. These iterations are also called generations ($g$), the size of the population (${P}_{size}$), a list of a 2-dimensional vector of the $x$ and $y$ coordinates of the nodes ($\textbf{L}$), a matrix of the Euclidean distances between the nodes ($\textbf{D}$), obtained using Equation \ref{eqSolveDistance}, for the distance between nodes $i$ and $j$, that contain the Cartesian coordinates $({c}^{i}_{x}, {c}^{i}_{y})$ and $({c}^{j}_{x}, {c}^{j}_{y})$, respectively:
\begin{equation} 
    \label{eqSolveDistance}
    \begin{split}
        d(\textbf{{c}}^{i}, \textbf{{c}}^{j}) = \sqrt{({c}^{i}_{x} - {c}^{j}_{x})^2 + ({c}^{i}_{x} - {c}^{j}_{y})^2} \:.
    \end{split}
\end{equation}

\begin{algorithm}[!ht]
    \SetKwInOut{Input}{Input}
    \SetKwInOut{Output}{Output}
    \SetKwFunction{Select}{Select}
    \SetKwFunction{Crossover}{Crossover}
    \SetKwFunction{Mutation}{Mutation}
    \SetKwFunction{EvaluateFitness}{EvaluateFitness}
    \Indm
    \Input{Size of population (${P}_{size}$); Number of elite individuals ($E$); Probability of mutation ($\gamma$); Number of iterations ($g$); Number of truck-drone pairs (${N}_{td}$); Number of nodes ($n$); Nodes location coordinates ($\textbf{L}$); Nodes distances matrix ($\textbf{D}$)
    }
    \Output{Candidate solution $\textbf{X}$}
    \Indp
        \BlankLine
        Generate random ${P}_{size}$ number of solutions and save them in $\textbf{P}$\;
        Evaluate the fitness of each solution in $\textbf{P}$\;
        Save the best solution in $\textbf{P}$ into $\textbf{X}$\;
        
        \BlankLine
        \For{$i = 1$ to $g$} {
            \BlankLine
            Select the best $E$ solutions in $\textbf{P}$ and save them in $\textbf{{P}}_{s}$\;
            Initialise ${\textbf{P}}_{n}$ as empty list\;
            
            \BlankLine
            \For{$j = 1$ to ${P}_{size}$} {
                \BlankLine
                Select a random solution in $\textbf{{P}}_{s}$ and save it in ${\textbf{X}}_{A}$\;
                Select a random solution in $\textbf{{P}}_{s} \notin {\textbf{X}}_{A}$ and save it in ${\textbf{X}}_{B}$\;
                Perform crossover of ${\textbf{X}}_{A}$ and ${\textbf{X}}_{B}$ and save it in ${\textbf{X}}_{C}$\;
                Perform mutation of ${\textbf{X}}_{C}$ based on $\gamma$ and save it in ${\textbf{X}}^{'}_{C}$\;
                
                \BlankLine
                \If {${\textbf{X}}^{'}_{C}$ is infeasible} {
                    Update ${\textbf{X}}^{'}_{C}$ with repaired ${\textbf{X}}^{'}_{C}$\;
                }
                
                \BlankLine
                Save ${\textbf{X}}^{'}_{C}$ into $\textbf{{P}}_{n}$
            }
            \BlankLine
            Evaluate the fitness of each solution in $\textbf{{P}}_{n}$\;
            
            \BlankLine
            \If {Fitness of $\textbf{X} <$ Fitness of best solution in $\textbf{{P}}_{n}$} {
                Update $\textbf{X}$ with best solution found in $\textbf{{P}}_{n}$\;
            }
            
            \BlankLine
            Update population list $\textbf{{P}}$ with the current generation of solutions in $\textbf{{P}}_{n}$\;
        }
     \Return $\textbf{X}$
     \caption{Evolutionary Algorithm to solve the VRPDi}
     \label{algoEAforVRPDi}
\end{algorithm}

The population's fitness in each iteration is evaluated by converting the genotype solution into a route to follow using \textit{genotype-phenotype mapping}, as described in \cite{holland1973genetic} and \cite{mathew2012genetic}. For minimisation problems, as is the case with the proposed EA in this paper, to generate non-negative values in all the cases and to reflect the relative fitness of individual solutions, it is necessary to map the underlying objective function of the problem to a fitness function form. One such transformation commonly adopted for fitness function mapping is to use the inverse of objective function value, thus resulting in $F(i) = \frac{1}{{z}(i)}$ to get the fitness value of $F(i)$. The transformation in the equation does not alter the location of the minimum. However, it converts a minimisation problem to an equivalent maximisation problem, where the greater the value of $F(i)$, the more chances that the solution has to be an optimal solution to the problem. In the equation above, ${z}(i)$ is the objective function value of the ${i}^{th}$ solution. This fitness procedure implementation is described in Algorithm \ref{algoEAPopFitness}.

\begin{algorithm}[h]
    \SetKwInOut{Input}{Input}
    \SetKwInOut{Output}{Output}
    \SetKwFunction{Sort}{Sort}
    \Indm
    \Input{List of population ($\textbf{P}$)}
    \Output{List of population with fitness values (${\textbf{P}}^{'}$)}
    \Indp
        \BlankLine
        Initialise ${\textbf{P}}^{'}$ with $\textbf{P}$\;
        
        \BlankLine
        \For {$i = 1 \text{ \textbf{to}  Length of } \textbf{P}$} {
            Update the Fitness of ${\textbf{P}}^{'}_{i}$ with $(1 / Total\:time \text{ of } {\textbf{P}}_{i})$\;
        }
        
        \BlankLine
         Sort ${\textbf{P}}^{'}$ by fitness in ascending order\;
    \BlankLine
    \Return ${\textbf{P}}^{'}$
    \caption{Fitness evaluation procedure}
    \label{algoEAPopFitness}
\end{algorithm}
The algorithm utilises the concept of elitism in selecting individual solutions for the next generation in the evolutionary process, as described in \cite{laumanns2000unified}. Algorithm \ref{algoEAPopSelect} describes this selection process. The primary purpose of using elitism in an EA is to keep the reference for promising areas of the search space across the generations. In practical terms, this enables the continuous exploitation of these promising areas, where the local or global optimum can be found. Elitism also ensures the best individual solutions are found, considering the entire processing of an EA in the last generation created, which would be the final solution.

\begin{algorithm}[h]
    \SetKwInOut{Input}{Input}
    \SetKwInOut{Output}{Output}
    \Indm
    \Input{List of population ($\textbf{P}$); Number of elite individuals ($\textbf{E}$)}
    \Output{List of individuals to be used for breeding ($\textbf{S}$)}
    \Indp
        \BlankLine
        Initialise ${\textbf{P}}^{'}$ with empty list\;
        Initialise $l$ with the length of $\textbf{P}$\;
        
        \BlankLine
        \For {$i = 1 \text{ \textbf{to} } \textbf{E}$} {
            Insert ${\textbf{P}}_{i}$ to the list ${\textbf{P}}^{'}$\;
        }
        
        \BlankLine
        \For {$i = 1 \text{ \textbf{to} } (l - \textbf{E})$} {
            Initialise $R$ with a random decimal value (between 0 and 100)\;
            Initialise $T$ with 0\;
            
            \BlankLine
            \For {$j = 1 \text{ \textbf{to} } l$} {
                $T$ equals to $\sum^{i}_{j=1} (\text{Fitness of } {\textbf{P}}_{j})$\;
                
                \BlankLine
                \If {$R < T$} {
                    Insert ${\textbf{P}}_{j}$ to the list ${\textbf{P}}^{'}$\;
                }
            }
        }
    \BlankLine
    \Return ${\textbf{P}}^{'}$
    \caption{Population selection procedure}
    \label{algoEAPopSelect}
\end{algorithm}

The crossover operator used by the algorithm is described in Algorithm \ref{algoEAPopCrossover}. As the EA regards the visited nodes sequence and the nodes delivery type as two different components of its genotype, as described in \cite{goldberg2014alleles}, a partially mapped crossover (PMX) based operator is adopted as the crossover operator in this algorithm implementation. The mutation operator adopted for this implementation is described in Algorithm \ref{algoEAPopMutation}.

\begin{algorithm}[h]
    \SetKwInOut{Input}{Input}
    \SetKwInOut{Output}{Output}
    \Indm
    \Input{Parent A (${\textbf{XA}}$); Parent B (${\textbf{XB}}$); Number of nodes ($n$);}
    \Output{Resulting child from the crossover operation ($\textbf{X}$)}
    \Indp
        \BlankLine
        Initialise $\textbf{X}$ with empty solution genotype\;
        Initialise ${\textbf{G}}_{A}$ with a random decimal value (between 0 and 1) $*$ $n$\;
        Initialise ${\textbf{G}}_{B}$ with a random decimal value (between 0 and 1) $*$ $n$\;
        
        \BlankLine
        \For {$i = min({\textbf{G}}_{a}, {\textbf{G}}_{b}) \text{ \textbf{to} } max({\textbf{G}}_{a}, {\textbf{G}}_{b})$} {
            Insert ${\textbf{XA}}_{i}$ bit into $X$\;
        }
        
        \BlankLine
        \For {$i = 1 \text{ \textbf{to} Length of } \textbf{XB})$} {
            \BlankLine
            \If {bit i $\in \textbf{XB}$ AND bit $i \notin X$} {
                Insert ${\textbf{XA}}_{i}$ bit into $\textbf{X}$\;
            }
        }
    \BlankLine
    \Return $\textbf{X}$
    \caption{Crossover procedure}
    \label{algoEAPopCrossover}
\end{algorithm}

\begin{algorithm}[h]
    \SetKwInOut{Input}{Input}
    \SetKwInOut{Output}{Output}
    \Indm
    \Input{Solution (${\textbf{X}}$); Change of mutation ($\gamma$); Number of nodes ($n$);}
    \Output{Mutated solution (${\textbf{X}}^{'}$)}
    \Indp
        \BlankLine
        Initialise ${\textbf{X}}^{'}$ with $\textbf{X}$\;
        
        \BlankLine
        \For {$i = 1 \text{ \textbf{to} } n$} {
            Initialise $R$ with a random decimal value (between 0 and 1)\;
            
            \BlankLine
            \If {$R < \gamma$} {
                Initialise $S$ with value of $i$\;
                
                \BlankLine
                \While{$S == i$} {
                    Update $S$ with a random decimal value (between 0 and 1) $*$ $n$\;
                }
                
                \BlankLine
                Initialise ${\textbf{G}}_{A}$ with ${\textbf{X}}^{i}_{i}$\;
                Initialise ${\textbf{G}}_{B}$ with ${\textbf{X}}^{i}_{S}$\;
                Update ${\textbf{X}}^{'}$ at index $S$ with ${\textbf{G§}}_{A}$\;
                Update ${\textbf{X}}^{'}$ at index $i$ ${\textbf{G}}_{B}$\;
            }
        }
    \BlankLine
    \Return ${\textbf{X}}^{'}$
    \caption{Mutation procedure}
    \label{algoEAPopMutation}
\end{algorithm}

The repair mechanism adopted for this implementation is described in Algorithm \ref{algoEAPopRepair} for instances where two consecutive drone nodes are scheduled, which does not conform to the VRPDi model constraints. The procedure was developed for repairing infeasible solutions based on what \cite{mitchell2003generepair}, \cite{fitzGerald2008genetic} and \cite{ernst2023framework} proposed. The repair mechanism changes the second consecutive drone node to a truck node without modifying the delivery sequence, meaning that after the repair procedure, the algorithm has found a solution different from the one it initially found. The resulting repaired solution may or may not already be part of the list of candidate individual solutions in that generation of the algorithm.

\begin{algorithm}[h]
    \SetKwInOut{Input}{Input}
    \SetKwInOut{Output}{Output}
    \Indm
    \Input{Solution ($\textbf{X}$)}
    \Output{Repaired solution (${\textbf{X}}^{'}$)}
    \Indp
    \BlankLine
        Initialise ${\textbf{X}}^{'}$ with $\textbf{X}$\;
        \BlankLine
        \For {$i = 1 \text{ \textbf{to} $($length of } \textbf{X}) - 1$} {
            \BlankLine
            \If {Delivery of ${\textbf{X}}^{'}_{i} == $ Drone AND Delivery of ${\textbf{X}}^{'}_{i + 1} == $ Drone} {
                Update Delivery of ${\textbf{X}}^{'}_{i}$ to Truck\;
            }
        }
    \BlankLine
    \Return ${\textbf{X}}^{'}$
    \caption{Solution repair procedure}
    \label{algoEAPopRepair}
\end{algorithm}

Equation \ref{eqNoTruck} is used to compute the number of truck-drone pairs (${N}_{td}$) for a given dataset $\textbf{S}$. Where ${\theta}({\textbf{S}})$ represents the total demand for the nodes in dataset ${\textbf{S}}$, the value of ${\theta}({\textbf{S}})$ is given by ${\theta}(\textbf{S}) = \sum_{{i} \in \textbf{{S}}} {\theta}_{i}$, for each node $i$ in the dataset $\textbf{{S}}$. $C$ denotes vehicle capacity. If the computed value of ${N}_{td}$ is not an integer, the value is rounded up to the nearest integer. After the ${N}_{td}$ is computed, each truck-drone pair is assigned several nodes to be delivered equal to or smaller than the vehicle capacity $C$.
\begin{equation} 
    \label{eqNoTruck}
    \begin{split}
        {N}_{td} (\textbf{{S}}) = \frac{{\theta}(\textbf{S})}{C} \: {N}_{td} \in \mathbb{Z}^{+}
    \end{split}
\end{equation}

\section{Results} \label{sectionResults}


The work presented in this paper investigated the feasibility of applying an evolutionary algorithm to solve the vehicle routing problem with drones with interceptions (VRPDi). The algorithm is evaluated on ten datasets from \cite{bouman2015datasets}. The selected datasets feature customer nodes that are uniformly distributed across the $x - y$ Cartesian plane. These datasets are summarised in Table \ref{tableBoumanDatasets}.

\begin{table}[h]
    \begin{center}
    \begin{tabular}{c|c}
    \hline 
        \bf Datasets & \bf Nodes \\ 
    \hline
        Uniform-71-n50 & 50 \\
        Uniform-72-n50 & 50 \\
        Uniform-73-n50 & 50 \\
        Uniform-91-n100 & 100 \\
        Uniform-92-n100 & 100 \\
        Uniform-93-n100 & 100 \\
        Uniform-1-n250 & 250 \\
        Uniform-2-n250 & 250 \\
        Uniform-5-n500 & 500 \\
        Uniform-6-n500 & 500 \\
    \hline
    \end{tabular}
    \end{center}
    \caption{Datasets for algorithm evaluation from \cite{bouman2015datasets}}
    \label{tableBoumanDatasets}
\end{table}

The algorithm parameters were set as follows:
\begin{itemize}
    \item Algorithm-specific parameters include the population size, the elitism rate, the mutation rate, and the maximum number of generations.
    \item Problem-specific parameters include the number of nodes $(n)$, the truck speed $({v}_{t})$, the drone speed $({v}_{d})$, the truck delivery time $({\omega}_{i})$, the drone delivery time $({\sigma}_{i})$ and the number of truck-drones pairs to be scheduled $({N}_{td})$. Table \ref{tableEAProblemParams} summarises the values assumed for these variables. The truck speed $({v}_{t})$ and the drone speed $({v}_{d})$ parameters for each of the problems are set based on values from \cite{bouman2015datasets}.
    \item Truck capacity $(C)$ was set to 40 for datasets with less than or equal to 100 nodes and 100 for datasets with more than 100 customer nodes. The dataset's demand for each node $i$ was set to $1$.
\end{itemize}

\begin{table}[h]
    \begin{center}
    \begin{tabular}{c|c}
    \hline 
        \bf Parameters & \bf Values \\ 
    \hline
        Crossover & PMX \\
        Elitism rate (\%) & 15 \\
        Encoding & Metameric \\
        Fitness function & Linear ranking \\
        Max. generations & 1000\\
        Mutation operation & Bit inversion\\
        Mutation rate (\%) & 30 \\
        Reinsertion & Fitness-based \\
        Population size & 150 \\
        Selection & Elitism \\
    \hline
    \end{tabular}
    \end{center}
    \caption{Algorithm parameters}
    \label{tableEAProblemParams}
\end{table}

\begin{table*}[h]
    \begin{center}
    \begin{adjustbox}{width=.9\textwidth}
        \begin{tabular}{r|c|c|c|c|c|c|c|c|c}
        \hline 
            \multirow{2}{*}{\textbf{Data}} & \multicolumn{3}{c|}{\textbf{VRP}} & \multicolumn{3}{c|}{\textbf{VRPDi}} & \multicolumn{3}{c}{\textbf{Performance}} \\
            \cline{2-10}
                & \textbf{Time} & \textbf{Distance} & \textbf{CPU} & \textbf{Time} & \textbf{Distance} & \textbf{CPU} & \textbf{Time} & \textbf{Distance} & \textbf{CPU} \\
        \hline
            Uniform-71-n50 & 58.61 & 712.31 & \textbf{5.02} & \textbf{37.63} & \textbf{662.34} & 7.53 & \textcolor[rgb]{.29,.68,.31}{55\%} & \textcolor[rgb]{.29,.68,.31}{7\%} & \textcolor[rgb]{.77,.12,.23}{-33\%} \\
            Uniform-72-n50 & 60.46 & 819.87 & \textbf{4.66} & \textbf{43.19} & \textbf{562.46} & 10.75 & \textcolor[rgb]{.29,.68,.31}{39\%} & \textcolor[rgb]{.29,.68,.31}{45\%} & \textcolor[rgb]{.77,.12,.23}{-56\%} \\
            Uniform-73-n50 & 58.40 & 770.00 & \textbf{5.13} & \textbf{39.03} & \textbf{509.72} & 8.00 & \textcolor[rgb]{.29,.68,.31}{49\%} & \textcolor[rgb]{.29,.68,.31}{51\%} & \textcolor[rgb]{.77,.12,.23}{-35\%} \\
            Uniform-91-n100 & 82.03 & 1683.95 & \textbf{10.34} & \textbf{52.75} & \textbf{1156.19} & 11.08 & \textcolor[rgb]{.29,.68,.31}{55\%} & \textcolor[rgb]{.29,.68,.31}{45\%} & \textcolor[rgb]{.77,.12,.23}{-6\%} \\
            Uniform-92-n100 & 80.44 & 1634.95 & \textbf{6.92} & \textbf{51.69} & \textbf{1107.40} & 12.40 & \textcolor[rgb]{.29,.68,.31}{55\%} & \textcolor[rgb]{.29,.68,.31}{47\%} & \textcolor[rgb]{.77,.12,.23}{-44\%} \\
            Uniform-93-n100 & 81.98 & 1681.81 & \textbf{7.13} & \textbf{51.17} & \textbf{1082.89 } & 11.72 & \textcolor[rgb]{.29,.68,.31}{60\%} & \textcolor[rgb]{.29,.68,.31}{55\%} & \textcolor[rgb]{.77,.12,.23}{-39\%} \\
            Uniform-1-n250 & 220.34 & 6566.14 & \textbf{27.86} & \textbf{144.00} & \textbf{3933.12} & 29.14 & \textcolor[rgb]{.29,.68,.31}{53\%} & \textcolor[rgb]{.29,.68,.31}{66\%} & \textcolor[rgb]{.77,.12,.23}{-4\%} \\
            Uniform-2-n250 & 224.39 & 6691.53 & 38.61 & \textbf{149.70} & \textbf{4013.30} &  \textbf{24.75} & \textcolor[rgb]{.29,.68,.31}{46\%} & \textcolor[rgb]{.29,.68,.31}{69\%} & \textcolor[rgb]{.29,.68,.31}{24\%} \\
            Uniform-5-n500 & 260.68 & 9730.62 & 149.37 & \textbf{154.23} & \textbf{6662.66} & \textbf{119.68} & \textcolor[rgb]{.29,.68,.31}{45\%} & \textcolor[rgb]{.29,.68,.31}{68\%} & \textcolor[rgb]{.29,.68,.31}{33\%} \\
            Uniform-6-n500 & 245.24 & 8842.61 & 168.39 & \textbf{149.23} & \textbf{6382.95} & \textbf{136.56} & \textcolor[rgb]{.29,.68,.31}{64\%} & \textcolor[rgb]{.29,.68,.31}{38\%} & \textcolor[rgb]{.29,.68,.31}{23\%} \\
        \hline
        \end{tabular}
    \end{adjustbox}
    \end{center}
    \caption{VRP vs VRPDi performance summary}
    \label{tableVRPvsVRPDiSystemPerformance}
\end{table*}

\subsection{Algorithm Benchmarking} \label{sectionEvoBenchmark}

The results of the algorithm developed in this paper were compared to the nearest neighbour heuristic for initial solutions self-adaptive neighbourhood search differential evolution (NNHis SaNSDE) algorithms in \cite{dillon2023investigating} and \cite{ernst2024phdthesis} for scheduling multiple truck and drone algorithms. The NNHis SaNSDE algorithm is a population-based metaheuristic EA. The algorithm was introduced in \cite{ernest2023dif}. The algorithm is built on the SaNSDE-based algorithm introduced in \cite{yang2008self}, combined with problem-specific nearest neighbour heuristics that introduce good solutions into the initial population to guide the search process. The delivery completion time for the VRPDi solutions of the algorithm in this paper against solutions from the NNHis SaNSDE I algorithm is summarised in Table \ref{tableEAvsNNHisSaNSDEIBenchmark}. The VRPDi solutions with a maximum drone distance constraint of this paper against those of the NNHis SaNSDE II algorithm, are summarised in Table \ref{tableEAvsNNHisSaNSDEIIBenchmark}. The algorithms were compared in datasets with three distribution types: double-centred, single-centred and uniformly distributed. This distribution refers to the placement of the node $x-y$ coordinates along the Cartesian plane.

\begin{table*}[h]
    \begin{center}
    \begin{adjustbox}{width=.7\textwidth}
        \begin{tabular}{r|c|c|c|c}
        \hline 
            \bf Data & \bf Nodes & \bf No. trucks / Clusters   & \bf EA & \bf NNHis SaNSDE I \\ 
        \hline
            Doublecenter-61-n20 & 20 & 2 & \textbf{40.09} & 57.29 \\
            Doublecenter-61-n20 & 20 & 2 & \textbf{40.09} & 57.29 \\
            Doublecenter-91-n100 & 100 & 2 & \textbf{91.78} & 94.52 \\
            Doublecenter-3-n250 & 250 & 2 & 188.205 & \textbf{166.12} \\
            Doublecenter-3-n375 & 375 & 2 & 262.71 & \textbf{214.52} \\
            Doublecenter-5-n500 & 500 & 2 & 309.49 & \textbf{240.86} \\
            Singlecenter-61-n20 & 20 & 4 & \textbf{15.13} & 24.61 \\
            Singlecenter-100-n100 & 100 & 4 & \textbf{31.78} & 36.84 \\
            Singlecenter-1-n250 & 250 & 4 & 88.055 & \textbf{59.46} \\
            Singlecenter-1-n375 & 375 & 4 & 92.49 & \textbf{62.17} \\
            Singlecenter-6-n500 & 500 & 5 & 117.95 & \textbf{77.49} \\
            Uniform-61-n20 & 20 & 4 & \textbf{10.76} & 15.80 \\
            Uniform-91-n100 & 100 & 4 & \textbf{47.66} & 48.08  \\
            Uniform-8-n250 & 250 & 4 & 65.32 & \textbf{48.09} \\
            Uniform-1-n375 & 375 & 4 & 71.68 & \textbf{57.43} \\
            Uniform-8-n500 & 500 & 4 & 77.54 & \textbf{66.62} \\
        \hline
        \end{tabular}
    \end{adjustbox}
    \end{center}
    \caption{VRPDi benchmarking of the EA vs NNHis SaNSDE I \cite{dillon2023investigating}}
    \label{tableEAvsNNHisSaNSDEIBenchmark}
\end{table*}

\begin{table*}[h]
    \begin{center}
    \begin{adjustbox}{width=.9\textwidth}
        \begin{tabular}{r|c|c|c|c|c|c}
        \hline 
            \bf Data & \bf Nodes & \bf Max. distance & \bf Max. drone distance & \bf No. trucks / clusters & \bf EA & \bf NNHis SaNSDE II \\ 
        \hline
            Doublecenter-71-n50 & 50 & 822.02 & 616.51 & 2 & \textbf{77.15} & 87.08 \\
            Doublecenter-91-n100 & 100 & 751.58 & 563.69 & 3 & \textbf{79.34} & 86.56\\
            Doublecenter-1-n250 & 250 & 772.27 & 579.20 & 3 &  228.43 & \textbf{132.71} \\
            Doublecenter-5-n500 & 500 & 896.52 & 672.39 & 5 & 221.60 & \textbf{162.01} \\
            Singlecenter-71-n50 & 50 & 327.35 & 245.52 & 2 & 51.51 & \textbf{35.22} \\
            Singlecenter-91-n100 & 100 & 451.54 & 338.66 & 3 & 81.01 & \textbf{65.63} \\
            Singlecenter-1-n250 & 250 & 478.34 & 358.75 & 3 & 161.91 & \textbf{97.43} \\
            Singlecenter-5-n500 & 500 & 546.21 & 409.66 & 5 & 181.81 & \textbf{101.70} \\
            Uniform-71-n50 & 50 & 249.41 & 187.06 & 2 & \textbf{36.70} & 38.17 \\
            Uniform-72-n50 & 50 & 256.03 & 192.02 & 2 & 40.96 & \textbf{39.78} \\
            Uniform-73-n50 & 50 & 253.04 & 189.78 & 2 & \textbf{40.12} & 41.48 \\
            Uniform-91-n100 & 100 & 263.07 & 197.30 & 3 & \textbf{43.70} & 46.84 \\
            Uniform-92-n100 & 100 & 270.17 & 202.62 & 3 & \textbf{40.09} & 40.66 \\
            Uniform-1-n250 & 250 & 266.60 & 199.95 & 3 & 80.38 & \textbf{67.56} \\
            Uniform-2-n250 & 250 & 259.27 & 194.45 & 3 & 79.09 & \textbf{54.52} \\
            Uniform-5-n500 & 500 & 276.50 & 207.37 & 5 & 87.43 & \textbf{55.38} \\
            Uniform-6-n500 & 500 & 74.57 & 205.93 & 5 & 79.31 & \textbf{52.49} \\
        \hline
        \end{tabular}
    \end{adjustbox}
    \end{center}
    \caption{VRPDi benchmarking of the EA vs NNHis SaNSDE II \cite{ernst2024phdthesis}}
    \label{tableEAvsNNHisSaNSDEIIBenchmark}
\end{table*}


\section{Discussion} \label{sectionDiscussion}

\begin{figure}[H]
    \centering
    \includegraphics[width=.4\textwidth]{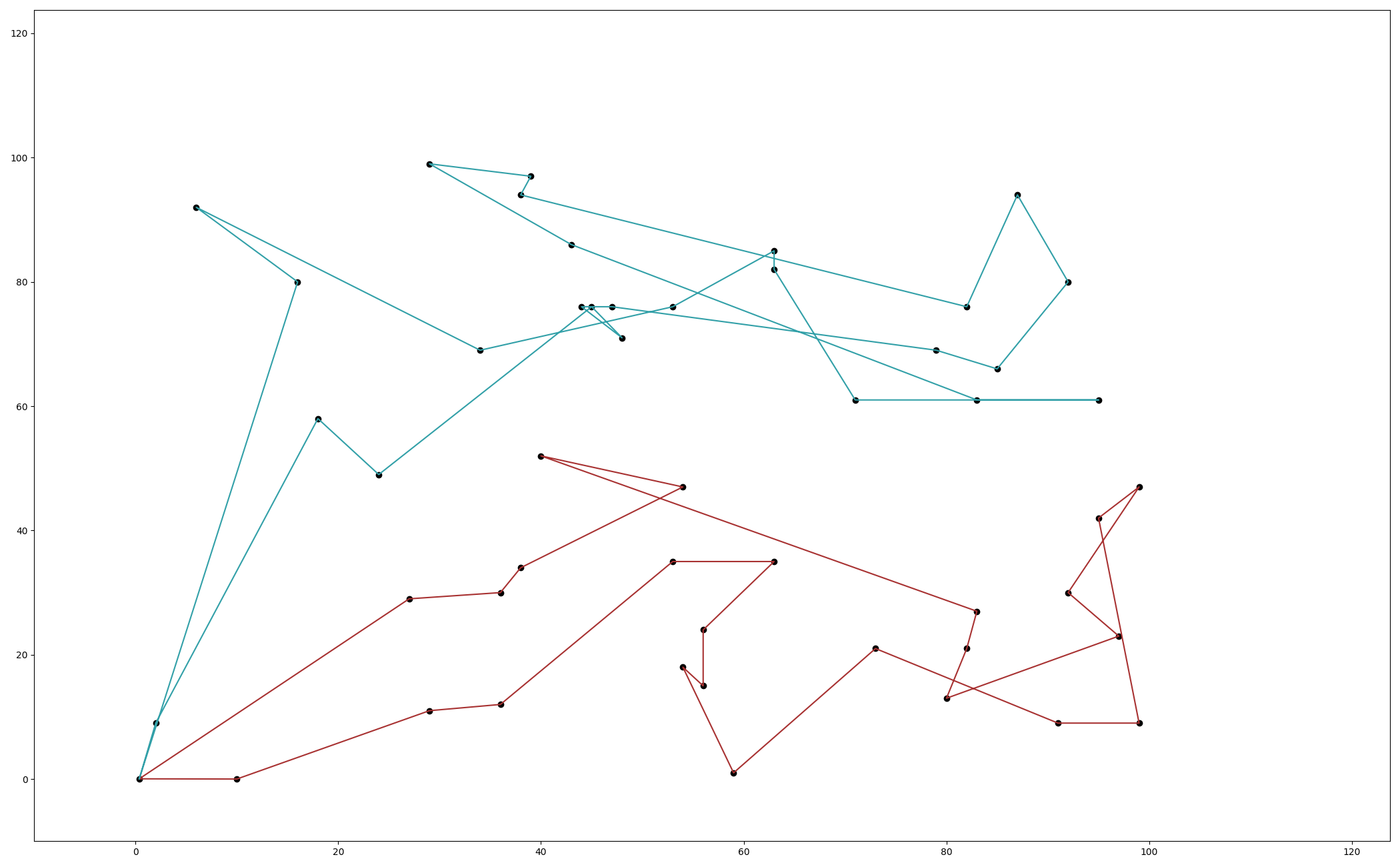}
    \caption{Deliveries for the VRP (Uniform-71-n50)}
    \label{figDeliveries50A}
\end{figure}

\begin{figure}[H]
    \centering
    \includegraphics[width=.4\textwidth]{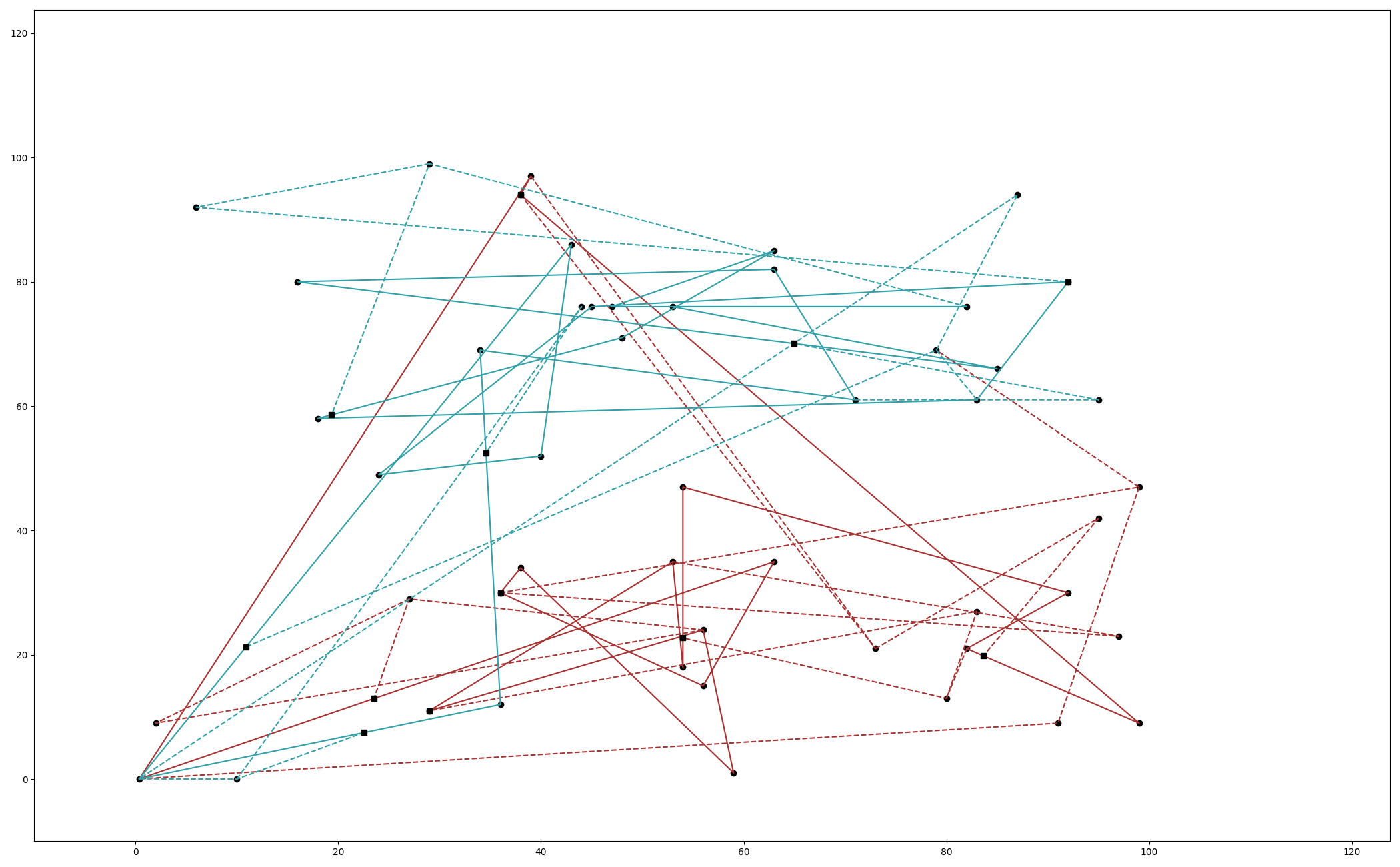}
    \caption{Deliveries for the VRPDi (Uniform-71-n50)}
    \label{figDeliveries50B}
\end{figure}

The algorithm is evaluated over 30 independent runs on the datasets. This section only contains graphical representations of \textit{Uniform-71-n50} dataset deliveries for the VRP and VRPDi. The delivery tours for each of the trucks of the VRP and the trucks and drones of the VRPDi for the dataset are shown in Figures \ref{figDeliveries50A} and \ref{figDeliveries50B}. The figure's continuous solid lines represent truck paths, the dotted lines represent drone paths, the black circles represent customer node locations, and the black squares represent interception points between trucks and drones. The above figure clearly shows that the solution obtained by the algorithm is feasible and complies with the problem constraints.

\begin{figure}[H]
    \centering
    \includegraphics[width=.45\textwidth]{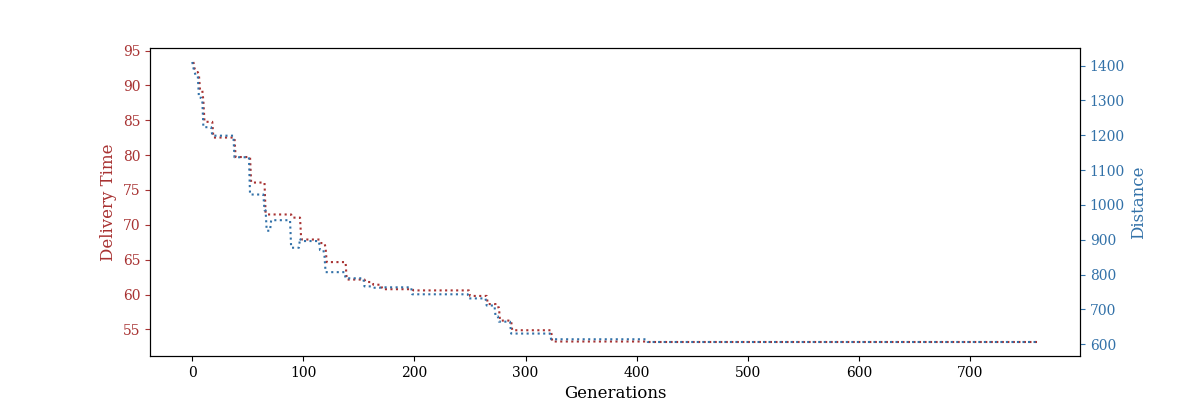}
    \caption{Best solutions for 50 nodes deliveries for VRP (Uniform-71-n50)}
    \label{figCost50A}
\end{figure}

\begin{figure}[H]
    \centering
    \includegraphics[width=.45\textwidth]{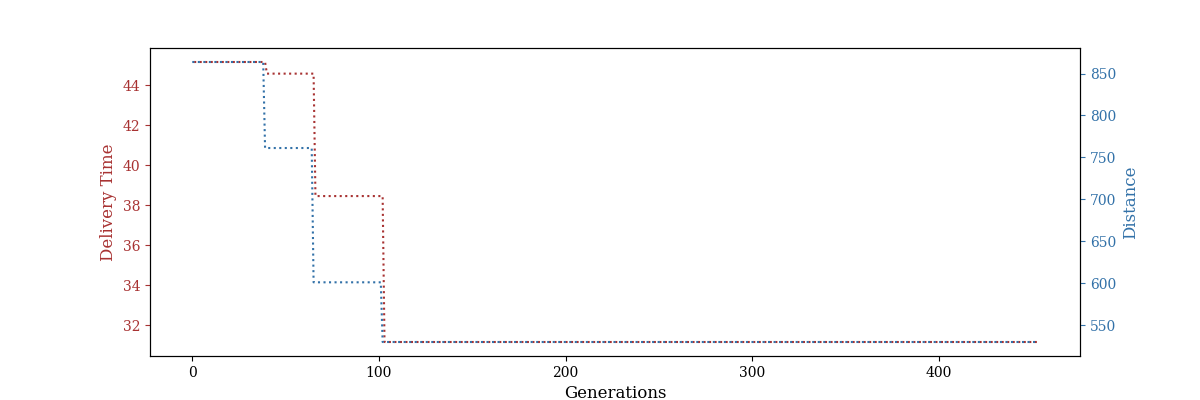}
    \caption{Best solutions for 50 nodes deliveries for VRPDi (Uniform-71-n50)}
    \label{figCost50B}
\end{figure}

The algorithm parameter settings balance exploration and exploitation throughout the search process. The algorithm's mean and standard deviations for the delivery time and the distance are shown in Table \ref{tableVRPvsVRPDiSystemPerformance}. \textit{Time} refers to the total time for completing all deliveries. \textit{Distance} is the total distance travelled by trucks. \textit{CPU} refers to the time in minutes the algorithm took to be completed for each of the runs. The VRPDi and VRP systems are compared using the same algorithm. Node deliveries are scheduled by a truck-drone assigning system, where nodes are randomly assigned to trucks and drones at first. As the algorithm progresses through the generations, nodes closer to each other are set using the Euclidean distance matrix mentioned in Section \ref{sectionApproach}, satisfying the problem's above-mentioned requirements. After that, the deliveries for the same truck-drone pair are scheduled truck-only.

Comparing the results of the VRP to those of the VRPDi, it is noticeable that the latter yielded positive results when compared with the VRP system in all datasets. There was an improvement in delivery time ranging between $39\%$ and $60\%$. Meanwhile, the distance improvements were between $7\%$ and $69\%$. The executions also registered a deterioration in the algorithm running time (CPU) as the number of nodes in the problems increased. This deterioration ranged between $4\%$ and $56\%$, with the most significant deterioration registered in the 50 and 100-node datasets. A complete summary of the results comparison of running the EA on the datasets in Table \ref{tableBoumanDatasets} for both VRP and VRPDi are shown in Table \ref{tableVRPvsVRPDiSystemPerformance}.

\begin{figure}[H] 
    \centering
    \includegraphics[width=.45\textwidth]{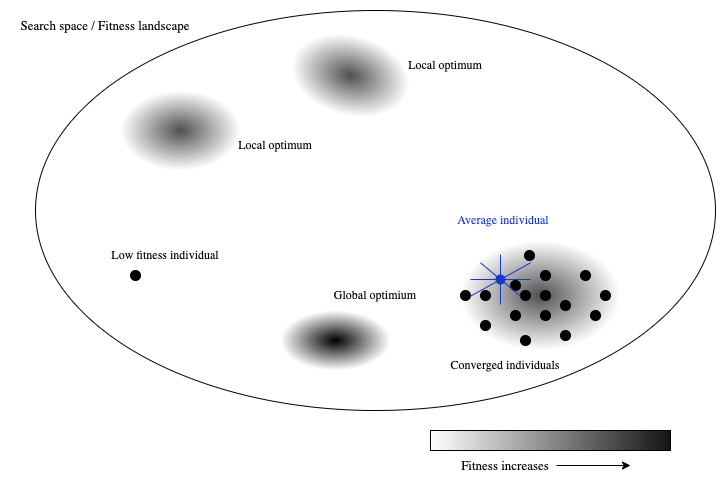}
    \caption{Search space depicting a converged population, global optimum, local optima, and the average individual's position \citep{mcginley2011maintaining}} 
    \label{figEAConvergedPopulation}
\end{figure}

\begin{figure}[H]
    \centering
    \includegraphics[width=.45\textwidth]{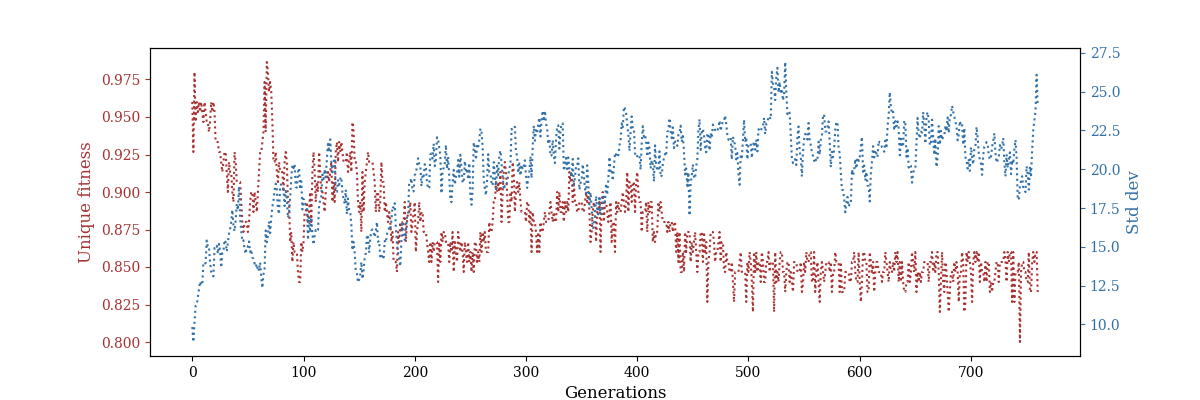}
    \caption{Solutions search diversity for 50 nodes for VRP (Uniform-71-n50)}
    \label{figDiversity50A}
\end{figure}

\begin{figure}[H]
    \includegraphics[width=.45\textwidth]{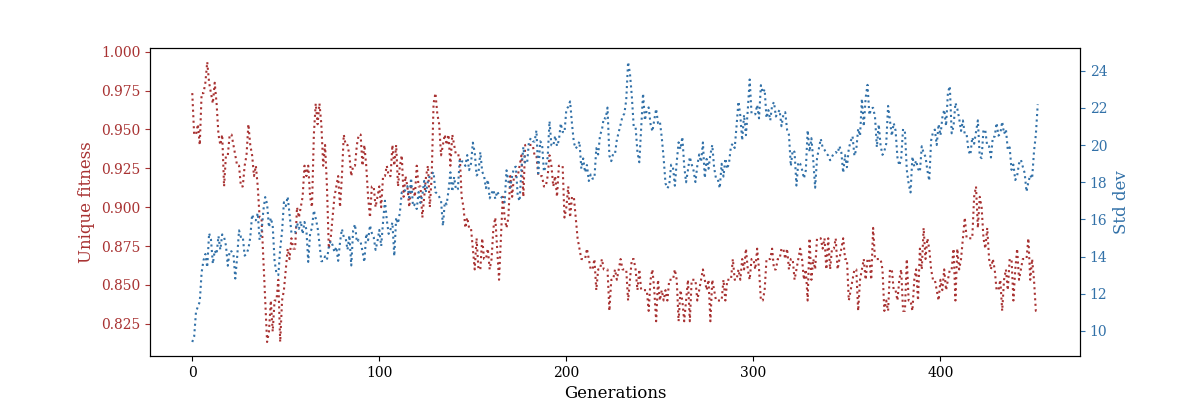}
    \caption{Solutions search diversity for 50 nodes for VRPDi (Uniform-71-n50)}
    \label{figDiversity50B}
\end{figure}

EAs often suffer from loss of diversity through premature convergence of the population, causing the search to be trapped in local optima \citep{zhu2003diversity, zhu2004population}. The phenotype measure describes the unique fitness values in a given population ($\textbf{P}$), divided by the population size (${P}_{size}$); the standard deviation of the fitness value is given by the equation $stddev(\textbf{P}) = \sqrt{\frac{ \sum^{N}_{i=1} ({f}_{i} - \bar{f})^{2} }{ N - 1}}$, over the generations of the algorithm. Genetic diversity represents a crucial part of evolutionary exploration since an EA can only search the space offered by the genes available in the population. Figure \ref{figEAConvergedPopulation} depicts a typical fitness environment that contains a converged population, the global optimum and several local optima. The location of the hypothetical average individual in the population is also illustrated in figure \citep{ursem2002diversity, zhu2004population}. The unique fitness value and the standard deviation value can be used to determine the diversity variation in the solutions in the algorithm search space. The above standard deviation equation can also be employed to determine the mutation and crossover rates of the EA. Thus, the search diversity shown in Figures \ref{figDiversity50A} and \ref{figDiversity50B} confirms that sufficient diversity is maintained for the search process and that the algorithm does not converge to a single solution early during the search.

Results of non-parametric statistical tests to evaluate the significance of results obtained from the algorithm executions are summarised in Table \ref{tableHyphotesisTest}. Each problem variation was compared to the effects of the other problem variation (VRP vs VRPDi) using a Mann-Whitney U test at $5\%$ significance. Wins, draws, and losses were recorded for all comparisons of the evaluated problem variation. A win is recorded if the first combination significantly outperforms the second. A draw is recorded if no statistical difference is observed. A loss is recorded for the first problem variation if the second variation outperforms it. These tests show that the VRP statistically underperformed compared to the VRPDi ten times when using delivery time as a performance metric, with no draws and ten losses recorded. Similar results were obtained when total distance was used as the performance metric for the evaluation.

\begin{table}[h]
    \begin{center}
    \small{
        \begin{tabular}{r|c|c|c|c|c|c}
        \hline 
            \multirow{2}{*}{} & \multicolumn{3}{c}{\bf Time} & \multicolumn{3}{|c}{\bf Distance} \\ 
        \cline{2-7}
            & \textbf{Win} & \textbf{Draw} & \textbf{Loss} & \textbf{Win} & \textbf{Draw} & \textbf{Loss} \\
        \hline
            VRP & 0 & 0 & \bf 10 & 0 & 0 & \bf 10 \\
            VRPDi & \bf 10 & 0 & 0 & \bf 10 & 0 & 0 \\
        \hline
        \end{tabular}
    }
    \end{center}
    \caption{Hypothesis tests regarding delivery time and distance of the VRP and the VRPDi}
    \label{tableHyphotesisTest}
\end{table}

\subsection{Benchmarking}

The algorithm in \cite{dillon2023investigating} (referred to as NNHis SaNSDE I algorithm), and the algorithm in \cite{ernst2024phdthesis} (referred to as NNHis SaNSDE II algorithm), use a cluster-first, route-second approach to solve the VRPDi. A maximum drone distance constraint is also used in the VRPDi variation in \cite{ernst2024phdthesis}. A drone can only travel a maximum distance equal to $75\%$ of the maximum distance between two nodes in the dataset. This means that if a drone travels from node $i$ to node $k$, to perform a delivery at node $k$, and the Euclidean distance between $i$ and $k$ is greater than the maximum drone distance of the dataset, the solution is deemed infeasible. Therefore, the repair procedure is triggered. This repair procedure involves changing the delivery type for node $k$ from a drone to a truck \citep{mitchell2003generepair, ernst2023framework}. For comparison purposes, the number of truck-drones pairs to be scheduled $({N}_{td})$ was set according to the number of clusters ($k$) for each of the problems being compared. These results show that the algorithm developed in this paper performed better than both NNHis SaNSDE algorithms for problems with $100$ or fewer nodes, regardless of node uniformity distribution. The algorithm developed in this paper did not outperform either of the algorithms in datasets with more than $100$ nodes.

\section{Conclusions} \label{sectionConclusions}

This paper proposes an evolutionary algorithm to solve the vehicle routing problem with drones with interceptions (VRPDi). The literature has shown the TSPD and the VRPD, including all its different variations, to be complex problems to solve, specifically as the number of nodes increases. This paper then discussed the different mathematical formulations of the VRPD, showing that solving the model proposed in this is more complex than unravelling a VRP. There is additional complexity in adding drones to assist the trucks in completing the deliveries. This paper demonstrated the performance of the proposed algorithm to solve such problems. The evaluation of the results of the proposed algorithm included solutions for the VRP (truck-only) and the VRPDi (truck-drone pairing) variations of the vehicle routing problem, which were done so that they could be compared. The paper also shows that the total delivery time improvement from a VRP to a VRPDi solution was between 39\% and 64\%. The improvement in total distance was between 7\% and 69\%, from VRP to VRPDi. The paper then compared the performance of the algorithm developed to the NNHis SaNSDE I algorithm from \cite{dillon2023investigating} and the NNHis SaNSDE II algorithm from \cite{ernst2024phdthesis}. 

The algorithm developed in this paper outperforms the NNHis SaNSDE I algorithm in datasets with $50$ and $100$ nodes. For the same datasets with more than $100$ nodes, the algorithm of this paper is outperformed by the NNHis SaNSDE I algorithm. The comparison also showed that the NNHis SaNSDE II algorithm outperformed the algorithm developed in this paper in datasets with more than $100$ nodes for the VRPDi problem variation with a maximum drone distance constraint. For the same problem variation, the algorithm developed in this paper outperformed the NNHis SaNSDE II algorithm in datasets with $100$ nodes or fewer nodes. Through the above benchmarking, it becomes evident that the algorithm in this paper performed better than the other algorithms in all variations of the VRPDi reviewed in this paper for $50$ and $100$-node datasets regarding the quality of the solutions and the computation effort required to arrive at the solution. It also became evident that the algorithm developed in this paper became more computationally expensive as the number of nodes in the problem increased, requiring an average of $128$ minutes to execute each run of the $500$ node dataset, highlighting the fact that its solution quality decreased as the number of nodes in the datasets increased. 

Given the increase in the demand for \textit{last-mile} deliveries from central logistic locations and their customers, enterprises aim to optimise these scheduling and routing activities to be as efficient as possible. The scheduling and routing algorithm developed from this paper can be utilised in any environment where deliveries must be scheduled from a central location.

The contribution of this work includes the development of the first GA-based algorithm to solve this multi-vehicle, namely trucks and drones, scheduling and routing problem, the VRPDi. The paper contributes to the metaheuristics optimisation field. These contributions include applying evolutionary algorithm concepts to solve scheduling and routing optimisation problems. The paper also contributes to evolutionary computing, namely, the genetic algorithm, which has been tested for solving this single-objective routing problem. Given the emergence of the fourth industrial revolution, companies operating in the transportation industry can benefit immensely from optimisation research into scheduling and routing optimisation. The VRPDi problem is characterised by a delivery system comprised of trucks and drones. An EA was developed to solve this scheduling and routing optimisation problem, leading to several areas being identified for future research. These opportunities are aimed at improving the approach used by the algorithm to solve the problem, thus improving the algorithm’s efficiency. In future work, we aim to utilise self-adaptive parameters in the algorithm to optimise control parameters for specific problem variations and increase the algorithm's efficiency by assigning nodes to trucks and drones using a cluster-first approach.

\subsection*{Abbreviations}

The following abbreviations are used in this manuscript: \\
\small{
    \begin{tabular}{@{}ll}
        CVRP & Capacitated vehicle routing problem \\
        EA & Evolutionary algorithm \\
        FSTSP & Flying sidekick travelling salesman problem \\
        GA & Genetic algorithm \\
        NNHis & Nearest neighbour heuristic for initial solutions \\
        PDSTSP & Parallel drone scheduling travelling salesman problem \\
        PMX & Partially mapped crossover \\
        SaNSDE & Self-adaptive neighbourhood search differential \\
        TSP & Travelling salesman problem \\
        VRP & Vehicle routing problem \\
        VRPDi & VRP with drones with interceptions \\
        VRPPD & VRP with pick-up and delivery \\
        VRPTW & VRP with time windows \\
    \end{tabular}
}
    

\subsection*{Author Contributions} 

Conceptualization: Pambo \& Grobler; Methodology: Pambo; Resources: Pambo \& Grobler; Writing - original draft: Pambo; Writing - review and editing: Pambo \& Grobler. 

\raggedright Both authors have read and agreed to the published version of the manuscript.

\bibliographystyle{styles/bib}

\end{document}